\documentclass{article}

\usepackage[nonatbib, preprint]{neurips_2021}

\usepackage[utf8]{inputenc} 
\usepackage[T1]{fontenc}    
\usepackage{hyperref}       
\usepackage{url}            
\usepackage{booktabs}       
\usepackage{amsfonts}       
\usepackage{nicefrac}       
\usepackage{microtype}      
\usepackage{xcolor}         
\usepackage{amsmath, amssymb}
\usepackage{graphicx}

\title{Deep Probabilistic Koopman: Long-term time-series forecasting under periodic uncertainties}

\author{
  
  Alex T.~Mallen\thanks{Code is available at \url{github.com/AlexTMallen/koopman-forecasting}.} \\ Department of Computer Science\\
  University of Washington\\
  Seattle, WA 98195\\
  \texttt{atmallen@uw.edu}

   \And
   Henning Lange \\
   Department of Applied Mathematics \\
   University of Washington \\
   Seattle, WA 98195 \\
   \texttt{helange@uw.edu} \\
   
   \AND
   J. Nathan Kutz \\
   Department of Applied Mathematics \\
   University of Washington \\
   Seattle, WA 98195 \\
   \texttt{kutz@uw.edu} \\

}

\begin{document}

\maketitle

\begin{abstract}
    Probabilistic forecasting of complex phenomena is paramount to various scientific disciplines and applications. Despite the generality and importance of the problem, general mathematical techniques that allow for stable long-term forecasts with calibrated uncertainty measures are lacking. For most time series models, the difficulty of obtaining accurate probabilistic future time step predictions increases with the prediction horizon. In this paper, we introduce a surprisingly simple approach that characterizes time-varying distributions and enables reasonably accurate predictions thousands of timesteps into the future. This technique, which we call Deep Probabilistic Koopman (DPK), is based on recent advances in linear Koopman operator theory, and does not require time stepping for future time predictions. Koopman models also tend to have a small parameter footprint (often less than 10,000 parameters). We demonstrate the long-term forecasting performance of these models on a diversity of domains, including electricity demand forecasting, atmospheric chemistry, and neuroscience. For electricity demand modeling, our domain-agnostic technique outperforms all of 177 domain-specific competitors in the most recent Global Energy Forecasting Competition.
\end{abstract}

\section{Introduction}
The goal of forecasting in scientific applications is to obtain models of complex phenomena that allow for accurate future-state predictions on unseen data. For many disciplines such as meteorology, finance, epidemiology, power systems, etc., long-term probabilistic predictions are paramount. In this paper, we leverage recent advances in operator theory, specifically Koopman theory~\cite{mezic2005spectral,Mezic2013arfm,brunton2021modern} to introduce a surprisingly simple class of models that allows for accurate predictions thousands of timesteps into the future. Koopman theory establishes that any non-linear dynamical system can be lifted, by the means of a non-linear functional, into a space (usually referred to as the observable space) in which its time evolution can be described by linear methods, in this case a linear dynamical system~\cite{koopman1931hamiltonian,koopman1932dynamical}. It can be understood as the time-dependent analogue to Cover's theorem, arguably the theoretical underpinning of deep learning~\cite{lecun2015deep} and Kernel techniques~\cite{scholkopf2001learning}. 

More formally, let $f(t)$ be a measurement of a non-linear dynamical system at discrete time $t$, then Koopman's theorem postulates that there always exists a linear operator $\mathcal{K}$ and a nonlinear functional $\psi$ such that $\mathcal{K} \psi(f(t)) = \psi(f(t+1))$. The linear operator $\mathcal{K}$ is usually referred to as the Koopman operator and might be infinite dimensional (e.g. in the case of a chaotic system $f$) whilst $\psi$ determines the observables. Algorithmic approaches that leverage Koopman's theorem usually assume a finite-dimensional Koopman operator (therefore excluding chaotic systems) and furthermore, in order to comply with the linearization constraint, restrict the scope to quasi-periodic phenomena. Note that the class of quasi-periodic systems is very large and includes many systems from the realms of the physical, biological and engineering sciences. Recent algorithms that leverage Koopman's theorem oftentimes solve a global optimization objective by the means of the Fast Fourier Transform \cite{lange2021fourier}, as opposed to Taylor-based approaches (including Neural approaches like Long-short Term Memory (LSTM) \cite{hochreiter1997long}) that solve a local objective by the means of a Taylor expansion (Gradient Descent). Koopman- or Fourier-based algorithms oftentimes exhibit a very small parameter footprint and do not require time-stepping for future state predictions.

In this paper, we generalize these techniques to the probabilistic setting by additionally making the assumption that uncertainty (or rather the parameters that describe a time-varying distribution) also exhibit periodic patterns. This simple but powerful assumption allows our model to extract cyclical patterns from time series data and ultimately produce competitive probabilistic forecasts of complex phenomena. We demonstrate the performance of the techniques on synthetic and real-world data sets from the realms of power systems, atmospheric chemistry and neuroscience.
\section{Related Work}

The techniques introduced in this paper are general in nature and therefore are, in some shape or form, related to time series models such as LSTMs~\cite{hochreiter1997long}, Gated Recurrent Units (GRUs)~\cite{chung2014empirical}, Neural and Gaussian processes~\cite{garnelo2018neural,garnelo2018conditional,williams1996gaussian} and convolutional time series models~\cite{bai2018empirical,sen2019think}. The reader is referred to \cite{hamilton2020time} for a recent review of different time series prediction techniques. Because our technique assumes notions of internal frequency and periodicity, Clockwork-RNNs are a closely related neuronal approach~\cite{koutnik2014clockwork}. There a similarities to the work of \cite{strangeGilpin} aimed at reconstructing nonlinear latent space dynamics to uncover intrinsic quasi-periodic structure and strange attractors. State space models (SSMs) are similar in character to our work as well, except that for SSMs the latent space itself is stochastic, which limits their long-term forecasting ability. Recent work on Deep State Space models has extended SSMs to incorporate RNNs as a model of the latent dynamics \cite{deepSSrangapuram}. Conceptually, this work could also be seen as a special case of (stochastic) NeuralODEs \cite{chen2018neural,li2020scalable} with linear latent dynamics. However, because of the linear latent dynamics, training is significantly easier and does not require the adjoint method or potentially expensive time steppers.

In the following, a brief overview of the history of Koopman theory and its algorithmic incarnations is given. The finding that any non-linear dynamical system can be linearized globally was first introduced by Koopman in 1931 for Hamiltonian systems~\cite{koopman1931hamiltonian} and was later generalized to continuous-spectrum systems \cite{koopman1932dynamical}. At the time, it was of considerable importance as a building block for advances in ergodic theory~\cite{birkhoff1931proof,neumann1932proof,birkhoff1932recent,neumann1932physical,moore2015ergodic}. Recently, interest in Koopman theory was renewed by work by \cite{mezic2005spectral,budivsic2012applied,Mezic2013arfm} and the development of {\em Dynamic Mode Decomposition}~\cite{schmid2010dynamic} (DMD) which provided a computational approximation to the Koopman operator~\cite{rowley2009spectral}. DMD was first introduced as a technique for modal decomposition to extract the dynamic behavior in fluid flows but has since been applied in various fields, such as neuroscience~\cite{brunton2016extracting}, epidemiology~\cite{proctor2015discovering}, acoustics~\cite{song2013global}, combustion modeling~\cite{moeck2013tomographic} and video processing~\cite{erichson2015compressed}. To be precise, DMD approximates the Koopman operator when the observables are restricted to direct measurements of the state variable~\cite{rowley2009spectral,Mezic2013arfm,Tu2014jcd,Brunton2016plosone,kutz2016dynamic,brunton2021modern}.

Many algorithms exist that attempt to estimate the Koopman operator from data. Most of these approaches rely on auto-encoder structures~\cite{otto2019linearly,lusch2018deep,yeung2019learning,wehmeyer2018time,takeishi2017learning} and solve an optimization objective that encourages linearity in `Koopman space' and prediction accuracy. These approaches have been extended in various ways, e.g., the authors of \cite{pan2019physics} extended Koopman theory to the probabilistic setting by employing Bayesian Neural Networks as encoders whereas in \cite{champion2019data}, the linearity constraint was relaxed and substituted for sparsity in the latent dynamics. These approaches usually do not consider linearity in `Koopman space' a constraint but rather the optimization objective and usually result in latent dynamics that are only approximately and locally linear.

In this paper, we build upon recent advancements that do away with the auto-encoder structure and enforce linearity and stability of the latent dynamics by construction~\cite{lange2021fourier,mendible2020data}. The resulting algorithm called Koopman Forecast will be discussed in more detail in the next section.

\section{Probabilistic Koopman Forecast}
\begin{figure}[t]
    \centering
    \includegraphics[width=0.9\linewidth]{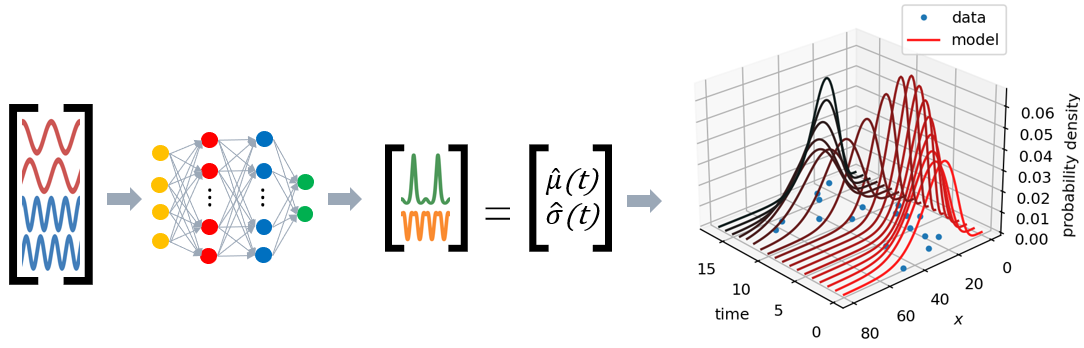}
    \caption{Left: an overview of our model when the data is assumed to be normally distributed. Right: our model recovers a probability distribution for every point in time given only a single realization from each distribution, and then projects into the future.}
    \label{fig:overview}
\end{figure}

As discussed earlier, the class of models introduced in this work build upon the Koopman Forecast (KF) algorithm~\cite{lange2021fourier}. The KF algorithm exploits the fact that the linear latent dynamics of any stable, non-linear dynamical system with finite dimensional discrete Koopman operator with unit magnitude complex eigenvalues can be expressed as a vector of sines and cosines. Fitting the KF algorithm to data points $x_t \in \mathbb R^n$ distributed uniformly over time therefore requires solving the following optimization objective:
\begin{align*}
    \mathcal{L}(\Theta, \vec{\omega}) = \sum_{t=1}^T \left( g_\Theta \left(\begin{bmatrix}
                \cos(\Vec{\omega} t)\\
                \sin(\Vec{\omega} t)
            \end{bmatrix}\right) - x_t \right)^2 = \sum_{t=1}^T L(\Theta,\vec{\omega},t)
\end{align*}
with $g$ usually being a NN describing the inverse of the observables $\psi$ with parameters $\Theta$. Even though this objective is non-convex and non-linear in $\vec{\omega}$, and therefore not amenable to optimization with stochastic gradient descent (SGD), it nevertheless can \emph{and needs to} be solved globally. This is achieved by exploiting periodicities in $L(\Theta,\vec{\omega},t)$, specifically that $L$ is periodic in $2\pi/t$ since $L(\Theta, \vec{\omega}, t) = L(\Theta, \vec{\omega} + 2\pi/t, t)$. This realization in conjunction with the Fast Fourier Transform (FFT) allows for the global reconstruction of $\mathcal{L}$ \cite{lange2021fourier}. Thus, the easiest way to describe KF is as a technique to fit a NN driven by sinusoids to data. Note that unlike $\Vec{\omega}$, $\Theta$ is optimized using SGD.

In this work, we extend this technique to allow for probabilistic forecasting for systems with periodically varying uncertainty. 
Specifically, we make the assumption that the data is drawn from a time-varying but, if conditioned on model parameters, independent random variable $X_t$ distributed as $P$ with parameters exhibiting quasi-periodic temporal patterns. We then task a DPK model to predict these parameters forward in time. Let $\theta_t$ denote the parameters of $P$, and $p(x_t|\theta_t) = P(X_t = x_t | \theta_t)$ denote $X_t$'s probability density or mass under $P$. For example, if we take $P$ to be a Gaussian, then $X_t \sim \mathcal{N}(\theta_t = \{\mu(t), \sigma(t)\})$. The assumptions of the respective algorithms can be summarized as:
\begin{align*}
   \underbrace{x_t = g_\Theta \left(\begin{bmatrix}
                \cos(\Vec{\omega} t)\\
                \sin(\Vec{\omega} t)
            \end{bmatrix}\right)}_{\text{Koopman Forecast}} ;\ \ \ \  
    \underbrace{x_t \sim P\left(X_t\ |\ \theta_t = g_\Theta \left(\begin{bmatrix}
                \cos(\Vec{\omega} t)\\
                \sin(\Vec{\omega} t)
            \end{bmatrix}\right)\right)}_{\text{DPK}}
\end{align*}
More concretely, we train the parameters \(\Theta\) of the NN \(g_\Theta\) to minimize the negative log likelihood of observing the data \(x_t\) given our model. In practice, a separate NN is used to fit each parameter.

Unlike KF, instead of minimizing the squared-error, we minimize the negative log-likelihood:
\begin{align*}
    \mathcal{L}_P(\Theta, \vec{\omega}) &= -\log \prod_{t=1}^T p(x_t | \theta_t) = - \sum_{t=1}^T \log p(x_t | \theta_t)\\
    &= \sum_{t=1}^T - \log p\left(x_t | g_\Theta \left(\begin{bmatrix}
                \cos(\Vec{\omega} t)\\
                \sin(\Vec{\omega} t)
            \end{bmatrix}\right)\right) = \sum_{t=1}^T L_P(\Theta, \Vec{\omega}, t)
\end{align*}
Note that independent of the choice of $P$, $ L_P(\Theta, \vec{\omega}, t) =  L_P(\Theta, \vec{\omega} + 2\pi/t, t)$ which in turn means that the `trick' of the KF algorithm to reconstruct $\mathcal{L}$ is applicable independent of the assumption of the underlying distribution. In general, our framework can be used as long as the density or mass function of the distribution $P$ is differentiable with respect to the distribution parameters.

By assuming that the parameters evolve according to a nonlinear function of sinusoids, we extract correlations between measurements that are a) adjacent in time because we assume $g$ to be smooth and b) multiples of one period apart. This constraint allows us to faithfully estimate a probability distribution for every point in time given only a single realization from each distribution, and then extrapolate that into the future as shown in Figure~\ref{fig:overview}.

In the simplest case, if we assume $P$ to be a time-varying Gaussian---i.e. \(X_t\sim\mathcal N\left(\mu(t), \sigma(t)\right)\), the loss becomes
\begin{equation}
    \mathcal{L}_\mathcal N = \sum_{t=1}^T\left(\frac{(x_t - \hat {\mathbf \mu}(t))^2}{2\hat{\mathbf \sigma}^2(t)} + \ln(\hat{\mathbf \sigma}(t))\right).
    \label{eq:normal_loss}
\end{equation}
This bears a strong resemblance to the ordinary mean-square error, except that squared errors are weighted inversely by \(\hat \sigma^2(t)\) and there is an added penalty for increasing \(\hat{\mathbf \sigma}(t)\). 


\section{Experiments}
\label{experiments}

All experiments are implemented in Python, making use of PyTorch \cite{pytorch} (BSD license), among other open source libraries such as \cite{vries2020allensdk, numpy, 2020SciPy-NMeth, pandas} (2-clause BSD license with non-commercial redistribution, BSD license, 3-clause new BSD license, new BSD license) and run on a single laptop CPU.
To optimize \(\mathcal L\) with respect to the NN parameters, we use SGD with a learning rate tuned to \(10^{-4}\) and a weight decay regularization of \(10^{-3}\). Data splits, loss functions, and any modifications to these parameters are noted in the respective experiment subsection. For most experiments, the data is normalized before passing through DPK, except in the case of discrete distributions and the synthetic experiments below.

The technique to extract angular frequencies \(\omega_i\) described in \cite{lange2021fourier}, because of the \emph{Unknown Phase Problem}, can oftentimes lead to numerical instabilities and therefore significant variance during training. To avoid this, for the natural data experiments, we hand-pick and fix \(\omega_i\). Thus, the natural data experiments probe the effectiveness of modeling complex phenomena by NNs driven by sinusoids rather than the KFs ability to extract these frequencies from data. Note that for the experiments described in this paper choosing \(\omega_i\) is trivial (365 days, 24 hours, ...).

\subsection{Synthetic experiments}
\subsubsection{Recovering time-varying distribution parameters: Gaussian and Gamma}
\label{toy}
We generate two sets of synthetic data to test our model's ability to recover probability distributions whose parameters evolve according to arbitrary non-linear but quasi-periodic functions of time. We conduct experiments on Gaussian- and gamma-distributed data. The gamma distribution, denoted \(\Gamma(\alpha, \beta)\), is typically used to model phenomena imbued with a non-negativity constraint and is parameterized by a shape \(\alpha\) and scale \(\beta\) variable. Both the Gaussian and the gamma model use a fully-connected NN with a 256-node first hidden layer and a 64-node second hidden layer with \emph{tanh} activation to model each distribution parameter independently.
For both data sets, we generate data for 100,000 time steps and recover the time-varying distribution parameters for the following non-linear functions:
\begin{align*}
    \text{Gaussian: } \mu(t) 
    &= 2\sin\left(1 + \sin\left(2\pi t/48 \right)\right); \quad
    \sigma(t) = \exp\left[\sin\left(2\pi t/31 \right) - 1\right] + 1/2 \\
    \text{Gamma: } \alpha(t) &= \left(\exp\left[\sin\left(2\pi t/96\right)\right] + \cos\left(2\pi t /12 \right)\right)^2 + 4;\\
    \beta(t) &= \sin\left(2\pi t/12\right)/2 + \cos\left(2\pi t/96\right) + 2
\end{align*}

\begin{figure}[t]
    \centering
    \includegraphics[width=0.9\linewidth]{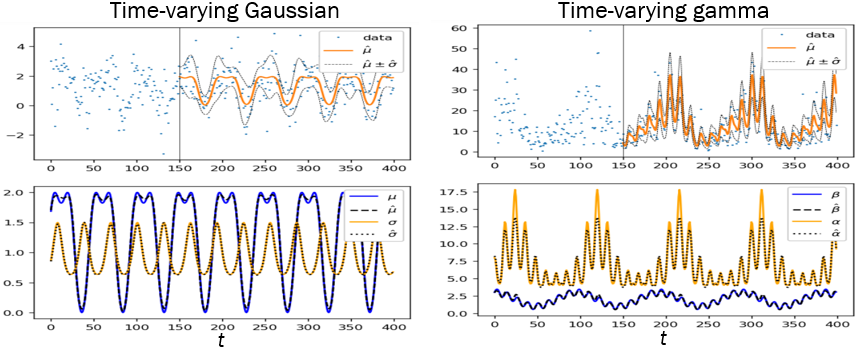}
    \caption{Top: Probabilistic forecasts of time-varying Gaussian- and gamma-distributed data. Bottom: model-learned parameters versus the parameters used to generate the data.}
    \label{fig:synthetic}
\end{figure}

Because parameters are perfectly quasi-periodic and overfitting is therefore less problematic, the weight decay is reduced to \(10^{-7}\) for this experiment. In both cases, the model converges quickly to the true parameters---in the case of the Gaussian after 100 epochs, and 1,000 epochs for the gamma-distributed data. Plots of the training data and recovered distribution parameters are shown in Figure \ref{fig:synthetic}.
 
\subsubsection{A probabilistic description of chaos: Duffing oscillator}
\label{duffing}

As discussed earlier, linearizing chaotic dynamical systems usually requires an infinite dimensional Koopman operator rendering algorithmic applications of Koopman theory difficult. Taking inspiration from ergodic theory and Liouville's theorem \cite{lutzen1985liouville}, in the following, we will demonstrate that the probabilistic description of dynamics introduced in this paper allows for a low-dimensional (in this case 1 dimensional) representation of chaos and in turn allows for the prediction of occupancy measures (as opposed to point forecasts). Because it is driven by a sinusoid, we choose the Duffing oscillator \cite{wiggins1987chaos} as an example.
The Duffing equation is a second-order differential equation that describes a family of oscillators varying from periodic to chaotic. It finds applications in modeling oscillators that do not obey Hooke's law \cite{rychlewski1984hooke} and is defined as:
\begin{align*}
    \ddot x + \delta \dot x + \alpha x + \beta x^3 = \gamma \cos(\omega t)
    \label{eq:duffing}
\end{align*}
We chose parameters in the chaotic regime, i.e. (\(\omega = 1.4, \gamma = 0.375, \delta = 0.1, \alpha = -1, \beta = 1\)), and discretize the resulting system spatially by choosing 20 equally sized bins between the minimum and maximum value $x_t$ takes. As the underlying probability distribution, we choose a categorical distribution with 20 categories representing the respective bin. The model is trained with a single driving frequency and a \emph{softmax} output with negative log-likelihood loss for 50 epochs with a learning rate of \(3\times10^{-4}\) on the first 100,000 time steps without regularization. The fully-connected NN uses a 128-node first hidden layer and a 256-node second hidden layer with \emph{tanh} activation.
\begin{figure}[t]
    \centering
    \includegraphics[width=0.97\linewidth]{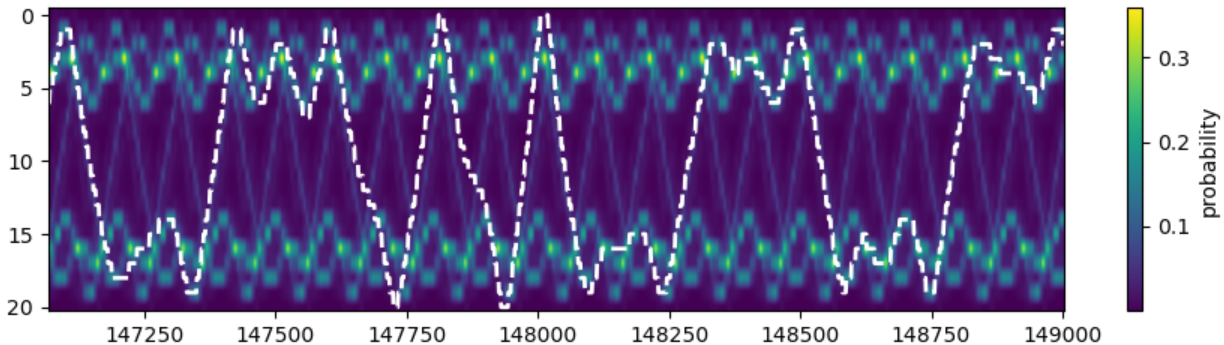}
    \caption{Probabilistic forecast of the chaotic Duffing oscillator. A brighter background indicates DPK believes the Duffing oscillator is more likely to occupy that region at that time.}
    \label{fig:duffing}
\end{figure}
Figure~\ref{fig:duffing} shows the predictions and a single realization of the system outside the training regime. The plot shows the probability that the system is in state $i$ (or bin $i$) at time $t$. DPK successfully extracts non-trivial occupancy patterns from the training data. For example, it discovered that the occupancy distribution is surprisingly sharp when transitioning between lobes, i.e. even though a point forecast of the systems behavior cannot be accurate past bifurcation points because of the systems chaotic nature, it is possible to predict the systems behavior when transitioning between lobes.
\subsection{Natural data experiments}
\subsubsection{Global Energy Forecasting Competition}
\label{gefcom}
Additionally, we evaluate our model on data from the most recent Global Energy Forecasting Competition (GEFCom) 2017. For each combination of month in 2017 and zone in New England, contestants are tasked with forecasting nine quantiles of the electricity demand for the respective month and zone at an hourly rate. The data is publicly available from ISO New England and can be found in \cite{HONG20191389}. 

Competing teams are evaluated relative to a competitive baseline model referred to as the vanilla benchmark (VB). VB is an ensemble of multiple linear regression (MLR) forecasts that account for temporal patterns and the effects of temperature \cite{HONG20191389}. Various temperature scenarios are drawn from the historical temperature data to create an ensemble of MLR forecasts, whose empirical distribution induces the uncertainty patterns in the model. 
Models are compared to VB using pinball loss:
\begin{equation}\text{Pinball}(\hat x, q, t) = \begin{cases} 
    (1 - \frac{q}{100})(\hat x_{q, t} - x_t) & \mbox{if }\hat x_{q, t} > x_t \\
    \qquad\ \ \frac{q}{100}(x_t - \hat x_{q, t}) & \mbox{otherwise} \end{cases},
\label{pinball}
\end{equation}
with $\hat x_{q, t}$ denoting the $q$th quantile of the respective probabilistic forecast \(\hat x\) at time $t$ and $x_t$ the observed value, where \(q \in \mathcal Q = \{10, 20,...,90\}\). Furthermore, let $E$ be the average of \(\text{Pinball}(\hat x, q, t)\), i.e.:
$E(\hat x) =  \frac{1}{9T} \sum_{t=1}^T \sum_{q\in \mathcal Q} \text{Pinball}(\hat x, q, t)$. The final score of each competing algorithm is the relative improvement over VB, i.e. \(R_\text{VB} = (1 - \frac{E}{E_{\text{VB}}})100\%\), with \(E_{\text{VB}}\) being the error of the vanilla benchmark. 

Because energy demand is driven by daily, weekly, and annual cycles, we choose and fix $\Vec \omega = [\frac{2 \pi}{24}, \frac{2 \pi}{7 \times 24}, \frac{2 \pi}{365.24 \times 24}]$. However, demand also has a non-periodic trend, which can be thought of as a limiting case as \(\omega_i \rightarrow 0^+\), so we also pass the time \(t\) since the start of training into the NN alongside the vector of sinusoids. We assume energy consumption to be skew-normally distributed. The skew-normal distribution, denoted \(\mathcal S(\xi, k, \alpha)\), is a generalization of the normal distribution that allows for nonzero skewness. Its PDF is given by the product of a normal PDF and CDF:

\begin{equation}
    f_{X \sim \mathcal S}(x; \xi, k, \alpha) = \frac{1}{k \sqrt{2\pi}} e^{-\frac{(x - \xi)^2}{2k^2}}\int_{-\infty}^{\alpha \left(\frac{x - \xi}{k}\right)}\frac{1}{\sqrt{2\pi}}e^{-\frac{z^2}{2}}dz
    \label{eq:skewpdf}
\end{equation}

Note that, in general, the mean and standard deviation of a skew-normal distribution are \textit{not} \(\xi\) and \(k\). The maximum likelihood loss function is
\begin{equation}
    \mathcal L_\mathcal S = \sum_{t=1}^T\left(\frac{(x_t - \hat {\mathbf \xi}(t))^2}{2\hat k^2(t)} + \ln(\hat k(t)) - \ln\left(\int_{-\infty}^{\hat \alpha(t) \left(\frac{x_t - \hat \xi(t)}{\hat k(t)}\right)}\frac{1}{\sqrt{2\pi}}e^{-\frac{z^2}{2}}dz\right)\right).
    \label{eq:skew_loss}
\end{equation}
A differentiable and numerically stable method for computing the log of the normal CDF can be challenging to find and strategies to avoid this are described in the appendix. The skew-normal model is comprised of 3 independent fully-connected NNs, each with a 256-node first hidden layer and a 64-node second hidden layer (except for the \(\alpha\) parameter's NN, which only has 32 second hidden layer nodes) with \emph{tanh} activation.

Competing teams were allowed to exploit historical demand, temperature, and humidity data since 2003 to forecast demand, and usually incorporate information such as indicators whether or not a day is a holiday. Our model does not use any information other than demand itself. Thus, all predictions are performed out of the auto-structure of demand alone. 
We train our model on the years of historical data and produce a 1 month forecast (with a 52 day gap between training and testing as mentioned in \cite{SMYL20191424}). We then compute the required 9 quantile forecasts for each zone and month. Our training loss function is modified to increase the weight of accurately modeling recent data and data from the time of year being tested, as detailed in the appendix. We train for 120 epochs.

Table~\ref{vanilla-comp} shows the average relative performance \(\bar R_\text{VB}\) of the best 7 models globally from the 177 teams competing in the qualifying round versus that of our DPK approach. With a mean relative improvement \(\bar R_\text{VB}\) of 15.4\%, DPK outperforms all competing teams. See the appendix for \(R_\text{VB}\) values at each zone and month. The standard deviation across the 108 zone-month pairs is 13.4\%. We also computed the standard deviation over 10 randomly initialized trials for three randomly selected forecasts, which was 4.33\% for Connecticut in December, 1.79\% for Northeast Massachusetts in July, and 1.33\% for Southeast Massachusetts in October. 
As described in \cite{HONG20191389}, competitors used quantile regression, MLR, gradient boosting machines (GBMs), NNs, quantile gradient boosted regression trees (QGBRTs), and others \cite{SMYL20191424,simplegood}. 

\begin{table}[t]
  \caption{Comparison against top 7 of 177 GEFCom 2017 teams}
  \label{vanilla-comp}
  \centering
  \begin{tabular}{l p{0.5\linewidth} c}
    \toprule
    Team     & Description     & \(\bar R_\text{VB}\) \\
    \midrule
    Tangent Works & \it{unknown/proprietary} & 12.9\% \\
    It Can Be Done & Quantile Regression, GBM & 11.9\% \\
    Orbuculum & Quantile Random Forest, NN, GBM \cite{SMYL20191424} & 11.5\% \\
    Black Analytics & MLR \cite{HONG20191389} & 9.1\% \\
    Dmlab & XGBoost, QGBRT, Generalized Additive Models & 8.4\% \\
    Simple But Good & Quantile Regression, MLR  \cite{simplegood}  & 7.1\% \\
    Rain Benchmark & MLR \cite{HONG20191389} & 5.7\% \\
    \midrule
     & \textbf{DPK}       & \bf{15.4\%}  \\
    \bottomrule
  \end{tabular}
\end{table}

\begin{figure}[t!]
    \centering
    \includegraphics[width=0.85\linewidth]{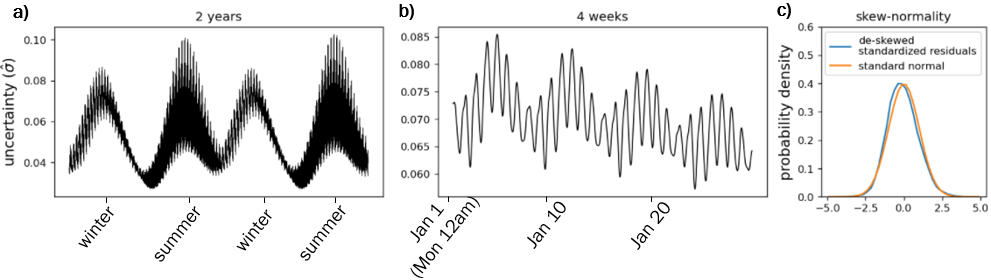}
    \caption{ Periodic trends in uncertainty over time. (a) and (b) demonstrate annual, weekly, and daily periodicities in uncertainty according to our model. (c) demonstrates that our model accurately captures its own uncertainty. Table~\ref{vanilla-comp} demonstrates that our model's uncertainty is representative of the uncertainty of the energy forecasting community as a whole.}
    \label{fig:periodicUncertainty}
\end{figure}

While DPK, as has been demonstrated, provides state-of-the-art forecasts based solely on temporal patterns, it also provides insightful descriptions of uncertainty. As can be seen in Figure~\ref{fig:periodicUncertainty}a and \ref{fig:periodicUncertainty}b, uncertainty is higher on weekdays compared to weekends, and in Summer and Winter than Spring and Fall, while the diurnal pattern depends on time of year. 

Due to time-varying uncertainty, errors can be evaluated by standardizing them using model parameters (i.e. model-standardized residuals \(r_t = (x_t - \hat \mu(t)) / \hat \sigma(t)\)). We measure the mean and root-mean-square (RMS) of the model-standardized residuals for each of the 108 forecasts as indicators of bias and overconfidence respectively, with 0 being perfect bias and 1 being perfect RMS. On the training interval, our model-standardized residuals had a mean bias of -0.003 and a RMS of 1.01, while they had a mean bias of -0.184 and a RMS of 1.19 for the test data. 
Month-by-zone mean bias and RMS can be found in the appendix. The distribution of de-skewed model-standardized residuals can be seen in figure~\ref{fig:periodicUncertainty}c. De-skewing involves mapping the quantiles of a skew-normal distribution to those of a normal distribution. Mathematically, \(z_{\text{de-skewed}} = \text{ICDF}_{\mathcal{N}_{0, 1}} (\text{CDF}_{\mathcal{S}_{\hat \xi, \hat k, \hat \alpha}}(x))\).

\subsubsection{Atmospheric pollution}
\label{atmochem}
\begin{figure}[t]
    \centering
    \includegraphics[width=0.9\linewidth]{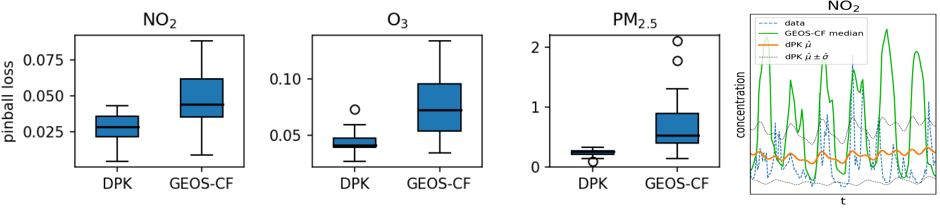}
    \caption{Comparison against GEOS-CF atmospheric chemistry forecasts for 3 pollutants and 50 stations.}
    \label{fig:NASApinball}
\end{figure}
Nitrogen dioxide (NO\(_2\)), ozone (O\(_3\)), and small particulate matter (PM\(_{2.5}\)) are three critical atmospheric pollutants. That is why the National Aeronautics and Space Administration (NASA) makes regular forecasts of their concentrations in more than 1000 global locations using the Goddard Earth Observing System Composition Forecast (GEOS-CF). The data is publicly available at \cite{kellerGEOS}. GEOS-CF uses a model physics package as described in \cite{kellerGEOS} to simulate the next 5 days of these pollutants everyday. Thus, there are 5 model forecasts for every hour of the day. Although performed at different times, in order to allow for probabilistic comparisons, we interpret the ensemble of these predictions as an empirical distribution.

For this comparison, we take 50 random locations, train our model on the available data from just 2018, and make a forecast for the entire year of 2019. The observations are recorded hourly, but there are many gaps of up to a few months, so we pass a vector of times \(t\) into DPK because \(x\) is not uniform over time. Because of the large amounts of missing data, the number of training epochs is dynamically set to \(\lfloor \frac{2,000,000}{\text{\# training datapoints}} \rfloor\). We assume that concentrations of these pollutants follow a gamma distribution because they are non-negative and near zero. The same gamma model is used as for experiment~\ref{toy}. Note that our forecast is an average of 6 months into the future, while the GEOS-CF forecast is an average of 2.5 days. A comparison of the pinball loss scores of DPK and the ensemble of GEOS-CF forecasts is shown in figure~\ref{fig:NASApinball}, showing DPK performs better and more consistently for all three pollutants. In contrast to the GEOS-CF predictions, DPK predictions show a bi-modal peak for NO\(_2\) suggesting that NO\(_2\) is affected by rush-hour traffic. Thus, DPK could find applications in validating physics-based modeling approaches that oftentimes rely on qualitative assumptions of e.g. pollutants emitted by commuter traffic. The appendix contains a more in-depth comparison that shows DPK outperforms GEOS-CF in terms of bias and root-mean-square error, and equals it in terms of Pearson correlation when evaluated as point forecasts. 

\subsubsection{Mouse cortical function}
\begin{figure}[t]
    \centering
    \includegraphics[width=0.8 \linewidth]{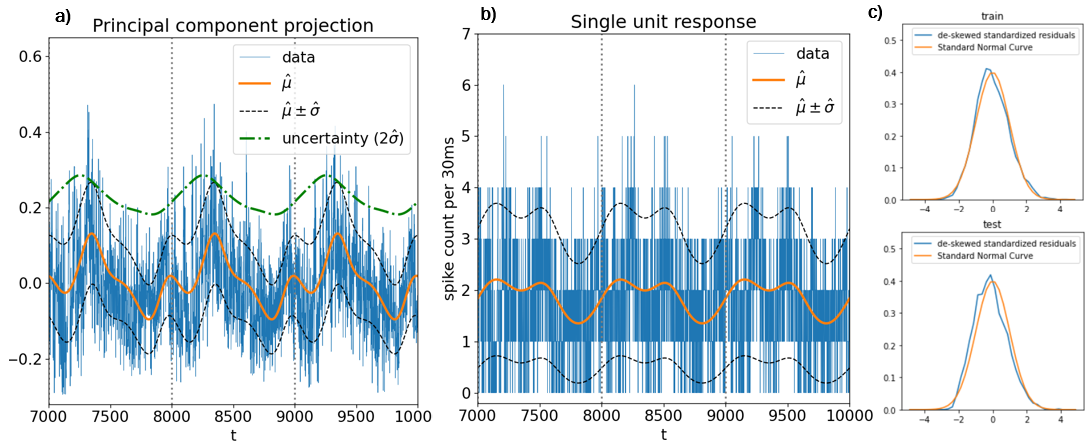}
    \caption{(a) A forecast of the principal component projection of a mouse's primary visual cortical response to 3 consecutive repetitions of a 30 second video. (b) A forecast of the response of an individual unit (neuron) to the video. (c) Distribution of de-skewed model-standardized residuals on train and test datasets for PCA projected data.}
    \label{fig:neural}
\end{figure}

With the goal of characterizing the strength and variability of a mouse's neural response to visual stimuli, we employed our model on recordings from a mouse's primary visual cortex in response to visual stimuli. The publicly available data was obtained from the Allen Brain Observatory visual coding dataset \cite{vries2020allensdk}. During the experiment, the mouse was presented with a 30 second natural movie looped 10 times.\footnote{\url{http://observatory.brain-map.org/visualcoding/stimulus/natural_movies}.} The neural data is therefore driven by a 30 second periodicity and this is the only frequency input to DPK. As a pre-processing step, we perform principal component analysis (PCA) on the firing data of the 60 measured units (one unit approximately corresponds to one neuron, barring measurement uncertainties) to extract correlated patterns of firing, and project the 60-dimensional neural firing data onto the first PCA mode. DPK is trained on the first 7 presentations of the video for 300 epochs with skew-normal loss, and makes a forecast for the remaining 3 loops. The hidden layers of the skew-normal model are the same as those of experiment~\ref{gefcom}. Neural firing is oftentimes described as a Poisson process for small time windows. Because of this, in a subsequent experiment, we model the spike counts of a particular unit using a time-varying Poisson random variable with rate \(\lambda\). The model is trained on the first 7 loops of the video for 600 epochs, and makes a forecast for the remaining 3 loops. The Poisson model uses a 256-node first hidden layer and a 64-node second hidden layer with \emph{tanh} activation.

Notice in figure~\ref{fig:neural}a that there are in fact strong periodic trends in uncertainty in the principal component projection of the response. On the training interval for PCA-projected data, our model-standardized residuals had a mean bias of -0.01 and a RMS of 1.00, while they had a mean bias of -0.152 and a RMS of 0.95 for the test data. During the first loop of the video, the neural response was greater because it was novel to the mouse, which likely accounts for the slight under-confidence of DPK.

\section{Conclusion}
\label{conclusion}
Many time series, while governed by complex dynamics, are ultimately driven by reliable periodic phenomena such as Earth's rotation and its position on the orbit around the sun. In this paper, we introduced a class of models that incorporates this simple but powerful assumption: that many phenomena can be characterized by a probability distribution whose parameters vary quasi-periodically with time. The resulting technique has a low computational and memory footprint and is easy to implement and understand. Despite the technique's simplicity, it oftentimes outperforms domain-specific competitors. We empirically show this on a range of challenging and important tasks including energy demand forecasting, atmospheric chemistry modeling, and neuroscience.

Furthermore, the methods introduced in this paper have a strong inductive bias. While this means they do not work well for data that is not quasi-periodic, the class of quasi-periodic phenomena is vast, and if the data is known to be produced by a  quasi-periodic system, an instantiation of a DPK model introduced in this paper might be the most appropriate modeling choice because unlike neural time series models, they will always produce quasi-periodic predictions. Thus, DPK constitutes a class of `stiff' models that can fit quasi-periodic phenomena tightly. While it is oftentimes hard to reason about what a neuronal time series model has learned, it is usually easy to understand what a DPK model 'does'. Even though it makes use of Neural Networks, a DPK model is easy to interpret because notions of periodicity persist. 

\textbf{Potential societal impact:} Just like any other Machine Learning algorithm, the techniques introduced here are imbued with an inductive bias. For many models such as complex Neural Network architectures, it is oftentimes difficult to reason about how this bias manifests itself. This is, for the most part, not true for the techniques introduced in this paper. It is easy to see that our model has a strong inductive bias towards quasi-periodicity and stability. Even if the model was fed with data that exhibits clear trends, its predictions will always be quasi-periodic and stable if the frequencies are finite. This might lead practitioners unaware of this property to wrong or potentially harmful conclusions. The models introduced in this paper cannot answer questions about whether or not a certain system is stable because stability is assumed. In the hands of a bad faith actor, the models predictions could be used to mislead or lie. A bad faith actor could use the model to predict carbon dioxide concentrations in the atmosphere and mislead people into believing that CO$_2$ concentrations do not exhibit a trend.
However, we believe that our model could have positive societal impacts because of its small computational footprint and its ease of use. The techniques introduced require little computational resources which leads to a small barrier of entry. The techniques also consume minimal electrical energy during training and inference.
\subsection{Future work}
\label{future_work}
Because this work builds upon the KF algorithm, it shares some of the same short-comings. The main draw back of the KF algorithm are the numerical instabilities when estimating optimal frequencies caused by the \emph{Unknown Phase Problem}. In its current inception, the KF algorithm is an interactive tool for modeling rather than a fire-and-forget approach. Oftentimes accurately finding the optimal frequencies requires running the algorithm from different initial conditions, observing which frequencies were extracted and worked well, then fixing the frequencies and training the NN to convergence. Mitigating the effects of the \emph{Unknown Phase Problem} could make the KF algorithm and therefore the approaches introduced in this paper significantly more robust and widely applicable.

Another route of potential future work is generalizing DPK to arbitrary
distributions by the application of Bayesian approaches \cite{blei2017variational,kingma2013auto} in conjunction with flexible and scalable approximate posterior distributions parameterized by NNs \cite{rezende2015variational,louizos2017multiplicative}. For such an approach, the Bayesian encoder would be driven by sinusoids and map to a probabilistic latent space from which samples would be drawn and fed into the decoder. Such an approach could in principle learn arbitrary probability distributions and not require a priori knowledge of the shape of the distributions to model.

\begin{ack}
The authors are grateful to Christoph Keller for providing the GEOS-CF model and observed data.  This work was supported by the U.S. Department of Energy, Office of Science, Office of Advanced Scientific Computing Research (ASCR) under Contract DE-AC02-06CH11347. Pacific Northwest National Laboratory is operated by Battelle for the DOE under Contract DE-AC05-76RL01830.


\end{ack}

\appendix

\section{Appendix}

\subsection{GEFCom additional details}

\textbf{Loss weighting:} In order to improve performance in our comparison of DPK against competitors in GEFCom 2017, we weighted the DPK training losses by recency and how close they were to the time of year being tested.
\begin{align}
    W_\text{time of year} &= 1 + 0.4 \cos\left(\frac{2 \pi}{1 \text{ year}}(t - \text{middle of testing period})\right) \\
        \label{eq:recency}
    W_\text{recency} &= \text{sigmoid}\left(\frac{t - 8.25 \text{ years}}{1.1 \text{ years}}\right) + 0.747 \\
    W &=  W_\text{time of year} \cdot W_\text{recency}
\end{align}
Equation~\ref{eq:recency} has a vertical shift to make the training-period-average of \(W\) equal 1.
\begin{figure}[h!]
    \centering
    \includegraphics[width=0.45\linewidth]{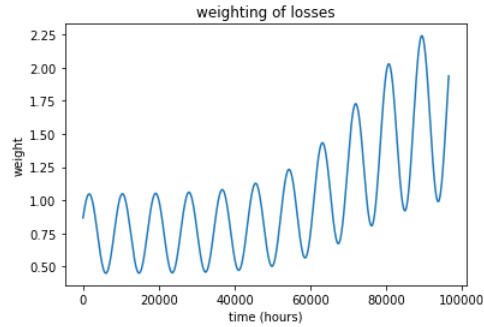}
    \caption{Training losses are weighted by recency and how close they are to the time of year being tested.}
    \label{fig:loss_weights}
\end{figure}

\textbf{Month-by-zone performance values} Running all 108 forecasts with the seed provided in the code (633) produces the following results.

\begin{figure}
    \centering
    \includegraphics[width=0.9\linewidth]{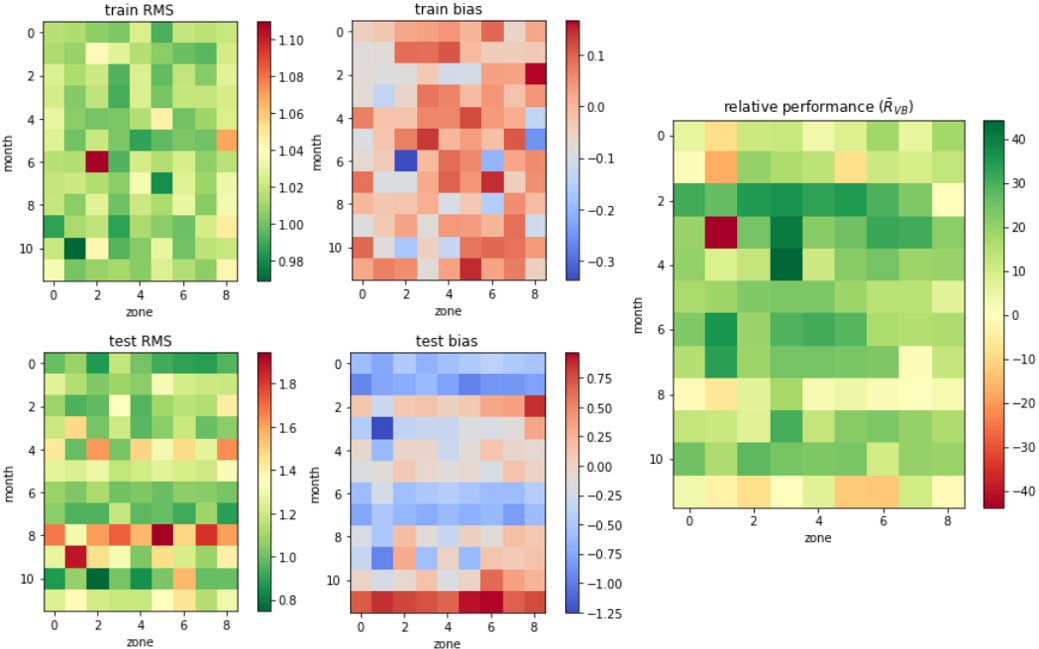}
    \caption{Performance visualization at each month and zone. The zones are, in order, ISONE CA, ME, RI, VT, CT, NH, SEMASS, WCMASS, and NEMASSBOST.}
    \label{fig:month-by-zone}
\end{figure}

Note that RMS and mean bias is calculated on de-skewed, model-standardized residuals.
\begin{equation}
    z_\text{de-skewed} = \text{ICDF}_{\mathcal N_{0,1}}(\text{CDF}_{\mathcal S_{\hat \xi, \hat k, \hat \alpha}}(x))
\end{equation}
Because the residuals are standardized, the mean should be 0 and the RMS should be 1. If the RMS is greater than 1, the model is overconfident and vice versa.

\begin{table}[b!]
    \caption{\(R_\text{VB}\) for each month and zone (\(\bar R_\text{VB}\))}
    \centering
    \begin{tabular}{lrrrrrrrrrr}
    \toprule
    {} &  ISONE &    ME &    RI &    VT &    CT &    NH &  SEMA &  WCMA &  NEMA &   \textit{avg} \\
    \midrule
    Jan &       5.8 &  -8.7 &  11.7 &  12.5 &   3.7 &   8.4 &    18.5 &     5.7 &        18.2 &   \textbf{8.4} \\
    Feb &       1.3 & -16.9 &  20.2 &  16.6 &  14.0 &  -7.9 &    11.2 &     9.6 &        13.2 &   \textbf{6.8} \\
    Mar &      30.9 &  26.9 &  34.6 &  36.4 &  33.0 &  35.0 &    29.4 &    22.9 &         0.2 &  \textbf{27.7} \\
    Apr &      20.1 & -43.6 &  24.7 &  41.1 &  22.3 &  25.3 &    31.9 &    31.1 &        21.7 &  \textbf{19.4} \\
    May &      20.6 &   8.6 &  12.9 &  44.5 &  11.2 &  22.4 &    24.7 &    19.1 &        19.9 &  \textbf{20.4} \\
    Jun &      17.4 &  19.0 &  22.8 &  23.7 &  23.7 &  19.3 &    14.5 &    14.6 &         7.0 &  \textbf{18.0} \\
    Jul &      23.0 &  36.2 &  19.6 &  29.3 &  30.9 &  28.4 &    16.8 &    15.4 &        16.6 &  \textbf{24.0} \\
    Aug &      15.6 &  33.7 &  19.4 &  24.8 &  24.7 &  24.0 &    23.4 &     1.3 &        12.8 &  \textbf{20.0} \\
    Sep &      -0.6 &  -6.2 &   6.7 &  17.9 &   2.3 &   4.4 &     1.7 &     1.0 &         2.6 &   \textbf{3.3} \\
    Oct &      13.3 &  11.9 &   7.5 &  30.6 &  13.2 &  22.4 &    23.6 &    20.2 &        16.3 &  \textbf{17.7} \\
    Nov &      25.5 &  16.5 &  27.9 &  24.3 &  23.4 &  23.9 &    10.5 &    20.2 &        19.5 &  \textbf{21.3} \\
    Dec &       3.8 &  -3.3 &  -8.4 &  -0.2 &   7.4 & -12.2 &   -12.4 &    10.7 &        -0.4 &  \textbf{-1.7} \\
    \bottomrule
    \end{tabular}
\end{table}

\textbf{Consistency:} 3 month-zone pairs were selected at random to evaluate the consistency and repeatability of our model. 

\begin{table}[t]
  \caption{\(R_\text{VB}\) consistency w.r.t. random initialization}
  \label{consistency}
  \centering
  \begin{tabular}{lrrr}
    \toprule
    Trial     & Dec CT     & Jul NEMA & Oct SEMA \\
    \midrule
    1 & 6.9 & 16.3 & 25.5 \\ 
    2 & 5.2 & 16.4 & 24.1 \\ 
    3 & 11.3 & 13.2 & 23.5 \\ 
    4 & 5.8 & 16.2 & 20.2 \\ 
    5 & 5.6 & 16.5 & 24.3 \\ 
    6 & -2.4 & 18.5 & 24.3 \\ 
    7 & 2.2 & 14.8 & 23.8 \\ 
    8 & 11.4 & 16.4 & 23.6 \\ 
    9 & 12.7 & 12.2 & 22.5 \\ 
    10 & 7.1 & 17.5 & 23.7 \\
    \midrule
    st. dev. & \textbf{4.33} & \textbf{1.79} & \textbf{1.33}  \\
    \bottomrule
  \end{tabular}
\end{table}
 
\subsection{Atmospheric chemistry additional details}
Data can be downloaded from NASA's website here.\footnote{\url{https://gmao.gsfc.nasa.gov/gmaoftp/geoscf/UW/samples_20210512/}} Comparisons on stations 478 and 621 are excluded because there were no observations from the last 9 months of the training year.

Skill scores are metrics used in the atmospheric chemistry forecasting literature. In \cite{kellerGEOS}, the authors evaluated their models using three skill scores: normalized mean bias (NMB) 
\[\text{NMB} = \frac{\sum_{t=1}^T (\hat x_t - x_t)}{\sum_{t=1}^T x_t},\]
normalized root mean square error (NRMSE)
\[\text{NRMSE} = \frac{\sqrt{\frac{1}{T}\sum_{t=1}^T (\hat x_t - x_t)^2}}{x_{0.95} - x_{0.05}},\]
and Pearson correlation. \(x\) is the observed data, \(\hat x\) is the prediction of it, and \(x_{0.05}\) and \(x_{0.95}\) are the fifth and ninety-fifth quantiles of the data respectively. These are evaluations of a point forecast, so in order to do this analysis we took the mean value of our probabilistic forecast and the mean of the 5 GEOS-CF forecasts.
\begin{figure}
    \centering
    \includegraphics[width=\linewidth]{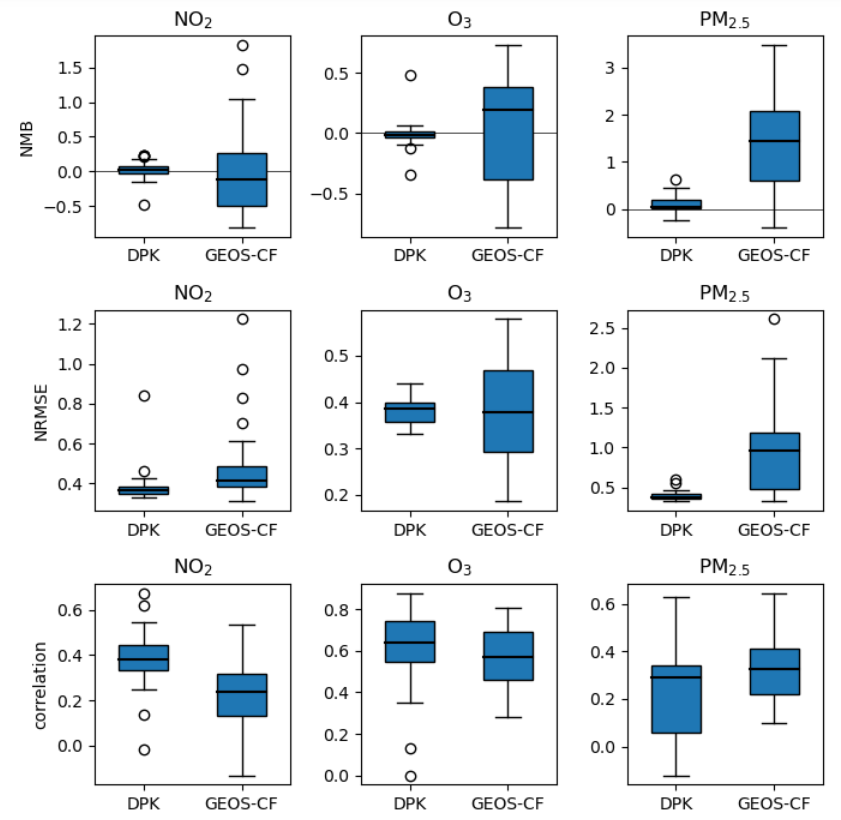}
    \caption{Skill score comparison}
    \label{fig:skill_scores}
\end{figure}

\subsection{Normal \emph{log-CDF} approximation}
Training DPK to find the time-varying parameters of a skew-normal distribution involves taking the log of the normal CDF in the maximum likelihood loss function (equation~\ref{eq:skew_loss}). When training data is many standard deviations below the model mean, the normal CDF is 0, and its logarithm is undefined. While such values are rare, they are common enough to necessitate a differentiable approximation to the standard \emph{log-CDF} for extremely negative values. For batches in which such data points exist, we employ the following piecewise approximation:
\begin{equation*}
    \text{logCDF}(z) \approx \begin{cases}
        \frac{-1}{2}z^2 - 4.8 + \frac{2509(z-13)}{(z - 40)^2(z-5)} & z < -0.1 \\
        \frac{1}{2}e^{-2z} - \frac{1}{5}e^{-(z - 0.2)^2} & \text{otherwise}
    \end{cases}
\end{equation*}

\begin{figure}
    \centering
    \includegraphics{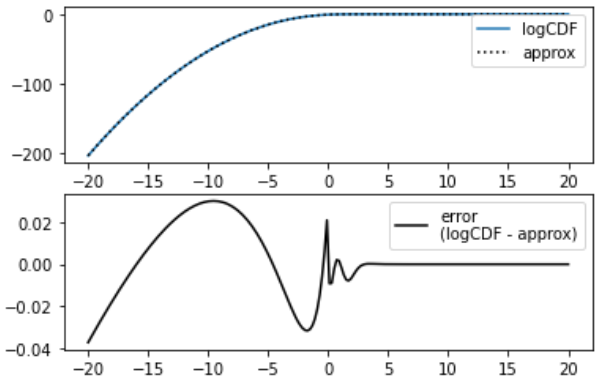}
    \caption{Differentiable approximation to logCDF for numerical stability. The approximation is compared to the non-differentiable \texttt{scipy.stats.norm.logcdf}~\cite{2020SciPy-NMeth}.}
    \label{fig:logcdf}
\end{figure}

The \(z < -0.1\) piece originates from~\cite{logcdfstack}. This approximation yields less than 0.04 error for all \(z>-20\), as shown in figure~\ref{fig:logcdf}. This approximation is not accurate enough to use for every batch, but typically less than 1\% of batches require it. One additional trick to mitigate this problem is to scale up the output of the \(\hat k\) NN by a constant (we use 10), so the initial standard deviation is large.

\subsection{Confidence training}

The confidence of our model is significantly affected by overfitting. If a DPK model overfits to the training data, it will be overconfident (RMS\(_{z_\text{de-skewed}}\) > 1). We therefore explore strategies to mitigate such overconfidence in this section.

One idea is to never train the neural network for two different parameters on the same training data, so that the parameters cannot `conspire' to overfit in a cooperative way. A related method for distributions such as the normal or perhaps skew-normal that have a parameter that effectively controls the variance is to train all parameters concurrently, except reserve some training data (which must critically never have been used to train the parameters controlling the mean of the distribution) to be used at the end to fine-tune the model's uncertainty. Both of these techniques require some method for partitioning the training data by parameter. While it may be tempting to choose a uniformly random partition, the continuity of real-world data makes it such that an overfit to one small portion of the training data will likely be a good model of adjacent data due to, for example, correlated temperature. In other words, short-term forecasts are easier than long-term forecasts. It is therefore advisable to make partitions as contiguous as possible (e.g. train \(\sigma\) on a hind-cast of the oldest training data) so as to replicate long-term forecasting. If there is \emph{any} periodic or trend-like structure to the partitioning, this will likely be exploited by the neural networks (e.g. a DPK model was able to reduce it's confidence only for the hind-cast portion of training by exploiting a long-period beating frequency between two similar frequencies). These techniques have yet to be rigorously tested for trade-offs between improved notion of confidence and smaller training data for each parameter.

\medskip

{\small
\bibliographystyle{plain}
\bibliography{references}
}

\end{document}